\documentclass[11pt,twocolumn]{article}
\usepackage{times}
\usepackage{latexsym}
\usepackage[T1]{fontenc}
\usepackage[utf8]{inputenc}
\usepackage{microtype}
\usepackage{inconsolata}
\usepackage{amsmath}
\usepackage{graphicx}
\usepackage{booktabs}
\usepackage{tabularx}
\usepackage{array}
\usepackage{xspace}
\usepackage[table]{xcolor}
\usepackage{placeins}
\usepackage{capt-of}
\usepackage{needspace}
\usepackage{enumitem}
\usepackage[a4paper,margin=2.5cm,heightrounded=true]{geometry}
\usepackage{natbib}
\usepackage{ebgaramond}

\usepackage{authblk}

\usepackage{footmisc}

\usepackage[colorlinks=true,linkcolor=blue!40!black,citecolor=blue!40!black,urlcolor=blue!40!black,bookmarksnumbered]{hyperref}
\hypersetup{
  pdftitle={SciPIvot: Benchmarking Principal-Investigator Judgment at Pivotal Moments of Scientific AutoResearch Agents},
  pdfauthor={Youran Sun and Xingyu Ren and Chugang Yi and Kejia Zhang and Di Wu and Yingkang Cao and Allen Zhou and Jianping Ye and Suying Liu and Jiaxuan Guo and Jianda Du and Weicheng Ye and Kecheng Liu and Hefan Zhang and Haixin Wang and Jie Xiao and Eric Yang and Haizhao Yang},
}

\newcommand{\bench}{Sci$\Pi$vot\xspace}

\newcommand{\epigraphblock}{%
  \par\begingroup
  \centering
  \vspace{0.55em}%
  {\color{black!55}\rule{0.3\textwidth}{0.5pt}}\par\nobreak
  \vspace{0.7em}%
  {\fontfamily{EBGaramond-OsF}\selectfont\itshape\large
    ``What I cannot create,\\
    I do not understand.''\par}%
  \nobreak\vspace{0.6em}%
  {\scshape\small\textls[150]{Richard Feynman}}\par
  \nobreak\vspace{0.3em}%
  {\itshape\footnotesize blackboard at Caltech, 1988}\par
  \nobreak\vspace{0.7em}%
  {\color{black!55}\rule{0.3\textwidth}{0.5pt}}\par
  \vspace{0.35em}
  \endgroup
}

\title{Sci$\Pi$vot: Benchmarking Principal-Investigator Judgment at Pivotal Moments of Scientific AutoResearch Agents}
\author[1*]{Youran Sun}
\author[2*]{Xingyu Ren}
\author[1]{Chugang Yi}
\author[1]{Kejia Zhang}
\author[1]{Di Wu}
\author[1]{Yingkang Cao}
\author[3]{Allen Zhou}
\author[1]{Jianping Ye}
\author[1]{Suying Liu}
\author[4]{Jiaxuan Guo}
\author[1]{Jianda Du}
\author[2]{Weicheng Ye}
\author[1]{Kecheng Liu}
\author[5]{Hefan Zhang}
\author[6]{Haixin Wang}
\author[3]{Jie Xiao}
\author[3$\dag$]{Eric Yang}
\author[1$\ddag$]{Haizhao Yang}
\affil[1]{University of Maryland, College Park}
\affil[2]{The Chinese University of Hong Kong}
\affil[3]{Gradient Network}
\affil[4]{Stanford University}
\affil[5]{Dartmouth College}
\affil[6]{University of California, Los Angeles}
\date{%
  {\small Code and data: \url{https://github.com/AutoResearch-Factory/SciPIvot}}\\[1.1em]
  \epigraphblock}

\begin{document}
\maketitle
\renewcommand{\thefootnote}{*}%
\footnotetext{Equal contribution.}
\renewcommand{\thefootnote}{$\dag$}%
\footnotetext{Corresponding author: \texttt{ey@gradientdata.ai}}
\renewcommand{\thefootnote}{$\ddag$}%
\footnotetext{Corresponding author: \texttt{hzyang@umd.edu}}
\renewcommand{\thefootnote}{\arabic{footnote}}

\begin{abstract}
% Terminology reference for the whole paper; authors only. One sentence per line. See CLAUDE.md.
\hyphenpenalty=1000 \exhyphenpenalty=1000 \tolerance=2000 \emergencystretch=2em
Large language model (LLM) agents can keep scientific experiments running for months with little human supervision, but reliable autonomous research also requires scientific judgment.
Evaluating scientific judgment is difficult because research decisions are open-ended and depend on project context.
We construct \bench, a 248-case benchmark of scientific judgment based on human corrections.
The benchmark draws on records from six months of research in which senior researchers ran LLM agents on 26 projects across six scientific fields and repeatedly intervened to correct the agents’ research judgments.
In each case, an agent chooses between the original LLM plan and the human-corrected alternative, using only the project record available before the correction as context.
We compare LLM and human performance and examine the effects of reasoning effort, access to the project record, and reconsideration.
At medium reasoning effort with the project record, models systematically prefer the original LLM plan over the human-corrected alternative.
Four of five frontier models perform below chance on the selected cases.
The highest accuracy in the project condition across the tested models and reasoning effort settings is 63.5\%, compared with 73.4\% for fresh researchers.
At medium effort, every model achieves lower accuracy with the project record than without it, and reconsideration does not consistently improve overall accuracy.
Even with a human-corrected alternative available, frontier models struggle with the selected research decisions, while \bench supports systematic evaluation of scientific judgment for LLM agents.

\end{abstract}

\tableofcontents

% One sentence per source line. Terminology, style rules, and figure placement: see CLAUDE.md.
\section{Introduction}
\label{sec:introduction}

Scientific research tackles problems with unknown answers.
Formulating useful questions, designing experiments, and revising assumptions as evidence arrives make scientific research a demanding test of general intelligence.
An AI system with this ability could study and improve its own models and methods.
Such progress could enable recursive self-improvement, as better systems conduct better research that helps create more capable systems~\citep{wijk2025rebench}.

We began with a practical question.
Can frontier models conduct high-quality scientific research autonomously?
Using frontier models with frontier agentic systems, we ran \emph{scientific AutoResearch agents} on original, PhD-level projects at scale across disciplines.
Over roughly half a year, individual projects ran for weeks to months with publication as the goal.
They involved theory, simulation, computation, and analysis of existing scientific data using CPUs and GPUs.
Keeping experiments running proved easier than choosing experiments that advanced the scientific question.
Automated checks and review still depended on models to notice mistaken assumptions and decide how to test them.
Failed judgments could stall projects or produce results that left the underlying problem unresolved.
Senior researchers had to diagnose the problem and steer the next experiment.

\begin{figure*}[t]
\centering
\includegraphics[width=\textwidth]{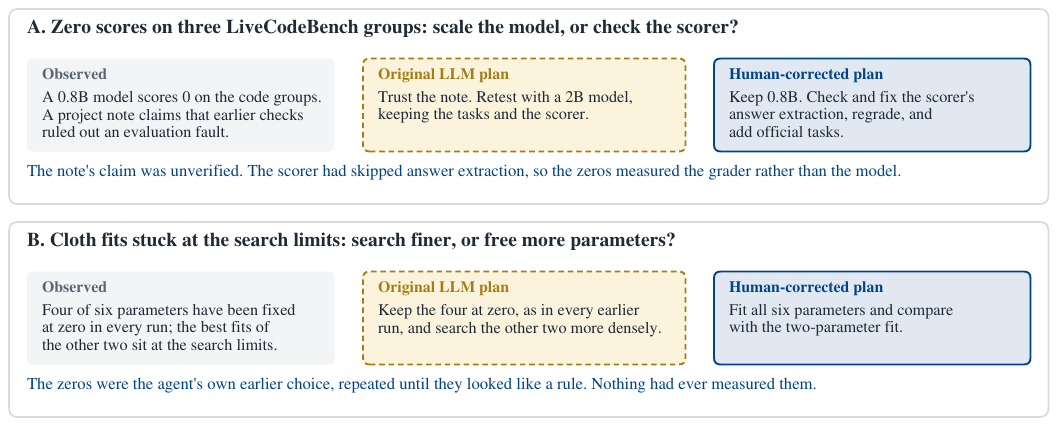}
\caption{Two human corrections from frontier research projects.
In (A), a small language model scores zero on three LiveCodeBench groups~\citep{jain2024livecodebench}, and the original LLM plan is to retest with a larger model.
The senior researcher retains the small model, and a subsequent audit finds that the scorer had skipped answer extraction.
In (B), a cloth fit reaches the search limits while four parameters stay fixed at zero, a restriction the agent had chosen in every earlier run.
The senior researcher asks to fit all six parameters.
Models receive the project record before the intervention and the two plans, without correctness labels or the explanations shown here.
Appendix~\ref{app:showcases} gives the source evidence and model responses.}
\label{fig:overview}
\end{figure*}

These interventions required anticipating what an experiment could establish.
Should an AutoResearch agent scale up, or first check its measurements?
Does a failed attempt undermine the idea, or only its implementation?
Such decisions repeatedly required human intervention (Figure~\ref{fig:overview}).
We call this scientific planning and decision-making \emph{Principal Investigator (PI) Judgment}.

% Figure 1 (the construction schematic) is input here so that it lands at the top of page 2; it is cited in Section 3.1.
\begin{figure*}[t]
\centering
\includegraphics[width=5.65in]{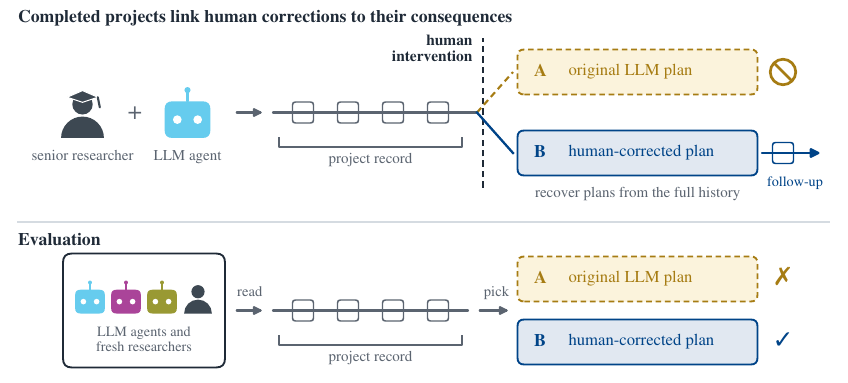}
\caption{How a case is built and used.
The top row shows an original LLM plan (A), a senior researcher's intervention, and the project's subsequent progress.
We use the completed history to reconstruct the human-corrected plan (B) and the evidence supporting it.
Evaluators receive the pre-intervention project record and both plans, then choose which to pursue.
Choosing B counts as correct.}
\label{fig:construction}
\end{figure*}

This dependence raises our next question.
How can we improve scientific AutoResearch agents so that fewer decisions require human intervention?
Agents can readily propose plausible plans, but executing an uninformative one can waste weeks of research.
The bottleneck is judging which plan merits those resources.
Human corrections expose these decisions and offer a way to teach them.

We introduce \bench to make these judgments measurable.
Its 248 cases from 26 projects pair original LLM plans with human-corrected alternatives reconstructed from completed project histories.
The model chooses between them using the project record available before the human intervention, including its code, experimental logs, and progress notes.

At medium reasoning effort with the project record, four of five models choose the original LLM plan more often than the human-corrected plan on this selected set.
Further deliberation can undo a correct choice and return to the same error, a failure we call \emph{re-trapping}.

\bench also connects to core NLP problems.
The case pairs provide preference supervision for scientific corrections rather than stylistic rewrites.
Scoring a choice evaluates the judgment of language agents, and the protocol exercises how agents use language to read a project record, compare two plans, and justify a choice.
These connections make \bench a testbed for NLP training and evaluation methods.

% One sentence per source line. Terminology, style rules, and figure placement: see CLAUDE.md.
\section{Scientific Judgment as a Bottleneck}
\label{sec:research-practice}

% Figure 2 (the two opening decisions) is input here so that it lands at the top of page 3; it is cited in the introduction and in Section 2.3.

\subsection{What We Require of AutoResearch}
Our AutoResearch program has three joint requirements.
\begin{enumerate}[leftmargin=*,topsep=3pt,itemsep=3pt,parsep=0pt]
\item \textbf{Autonomy.}
AutoResearch agents should work around the clock with minimal human intervention.
\item \textbf{Scale and scope.}
Many projects should run in parallel across scientific disciplines.
\item \textbf{Scientific quality.}
Projects should pursue original, PhD-level frontier scientific research worthy of publication.
\end{enumerate}
Frequent human intervention limits how many projects a human scientist can supervise, so autonomy and scale depend on better scientific judgment.

\subsection{What Scientific Quality Depends On}

Scientific quality depends on three contributors.
\begin{enumerate}[leftmargin=*,topsep=3pt,itemsep=3pt,parsep=0pt]
\item \textbf{Model capability.}
Models must interpret evidence, adapt to experimental results, and choose useful next steps.
\item \textbf{The agentic system.}
Systems must organize experiments, preserve project knowledge, and support review across long runs.
\item \textbf{Human scientific judgment.}
Human scientists currently diagnose scientific errors and steer experiments toward the research question.
\end{enumerate}

Agentic systems could reduce dependence on human judgment by adding rules and automated checks for observed failures.
However, a growing set of instructions becomes harder for models to follow reliably.
Extra instructions also consume tokens and can impair task performance.
Controlled report-writing experiments show declining adherence and, for some models, declining report quality as requirements accumulate~\citep{jaroslawicz2025how}.

Rules also leave scientific questions for the model to resolve.
An instruction to check a measurement does not specify which assumption is wrong or what evidence would expose it.
In our projects, existing checks and review did not reliably supply these judgments.
Senior researchers reviewed projects several times daily to identify stalled or misdirected work and choose the next test.
Without these corrections, continued execution could consume computation without resolving the scientific problem.
A capable AutoResearch agent must learn to supply this diagnosis and change of direction on its own.

\subsection{Two Examples of Human Scientific Judgment}
Figure~\ref{fig:overview} shows how a human correction changes what a project can learn from its next experiment.
\paragraph{Check the measurement before scaling up.}
The Transformer project asks whether reusing a pretrained model's layers, with small trainable adapters, can improve performance.
It first needs tasks on which the small base model can be evaluated meaningfully.
After observing zero scores on three LiveCodeBench groups, the agent proposes moving to a larger model.
The senior researcher keeps the small model and asks the agent to find suitable tasks.
A later audit finds that the scorer failed to extract answers from generated text.
Changing model size with the same scorer would leave this error in place.

\paragraph{Question the fit before refining the search.}
The cloth project asks how simulation parameters should change with mesh resolution to preserve realistic motion.
Its agent fixes four of six parameters at zero, a restriction carried forward from its own earlier runs, and proposes refining the two-parameter search after the best sampled fits reach the boundary.
The senior researcher points out that missing fitted values do not justify setting the corresponding physical parameters to zero and calls for fitting all six with the restricted fit as a comparison.

Both workflows could keep running without testing the mistaken assumptions that the corrections address.
We evaluate whether models recognize these scientific decisions from the project record.

% One sentence per source line. Terminology, style rules, and figure placement: see CLAUDE.md.
\section{From Human Corrections to Evaluation}
\label{sec:construction}
% Input at the top of Section 3 so this block lands on page 4; a table* appears on the page after its input point. See CLAUDE.md.
% Two tables side by side, then the two-panel radar figure with one shared caption.
\begin{table*}[!t]
\centering
\begin{minipage}[t]{.485\textwidth}
\vspace{0pt}
% GENERATED by scripts/build_paper_assets.py. Mirror any text change in that script, or a regeneration overwrites it. See CLAUDE.md.
% Generated from submission/data/project-coverage.csv by scripts/build_paper_assets.py.
% Table body; tables/coverage-profiles.tex supplies the float and width.
\small
\setlength{\tabcolsep}{3pt}
\renewcommand{\arraystretch}{1.08}
\begin{tabularx}{\linewidth}{@{}>{\raggedright\arraybackslash}Xrr@{}}
\toprule
Scientific field and examples & Projects & Cases \\
\midrule
\textbf{Applied mathematics \& scientific computing}\newline PDE solving; equation recovery; constraint calibration & 8 & 85 \\
\textbf{Robotics \& control}\newline Humanoid motion; programmatic control; navigation & 4 & 58 \\
\textbf{AI \& machine learning}\newline Language models; efficient training; scaling laws & 9 & 59 \\
\textbf{Biological imaging}\newline Cell tracking and lineages in 3D microscopy & 1 & 21 \\
\textbf{Mechanics \& materials}\newline Cloth simulation; material-property descriptors & 2 & 20 \\
\textbf{Earth sciences}\newline Rainfall downscaling; wellbore geology & 2 & 5 \\
\midrule
Total & 26 & 248 \\
\bottomrule
\end{tabularx}
\caption{Scientific fields represented in the benchmark, grouped by each project's primary objective.
The construction procedure does not depend on these area labels.}
\label{tab:project-coverage}

\end{minipage}
\hfill
\begin{minipage}[t]{.485\textwidth}
\vspace{0pt}
% Judgment categories; tables/coverage-profiles.tex supplies the float and width.
\small
\begin{itemize}[label={},leftmargin=0pt,labelsep=0pt,topsep=0pt,itemsep=4pt,parsep=0pt]
\item \textbf{Planning} (123 cases).
The agent starts costly runs before basic checks, or runs experiments too small to answer the question.
\item \textbf{Adapting} (65 cases).
It keeps the same hypothesis or search strategy even when earlier experiments have shown what should change.
\item \textbf{Anomaly awareness} (88 cases).
It overlooks suspicious results that should trigger checks for bugs, faulty measurements, or confounding factors.
\item \textbf{Evidence sufficiency} (47 cases).
It declares an idea a success or a failure before the experiments justify that conclusion.
\item \textbf{Repetition} (4 cases).
It repeats an experiment or computation whose answer is already available.
\item \textbf{Focus} (122 cases).
It tests a different claim or optimizes a proxy, leaving the project's actual research question unanswered.
\end{itemize}
\caption{Judgment categories, the number of cases involving each, and what the agent does wrong.
Counts include primary and secondary judgment labels across all 248 cases.
Categories can overlap.
Appendix~\ref{app:failure-cases} distinguishes judgment labels from historical-error coding.}
\label{tab:failure-taxonomy}

\end{minipage}
\par\medskip
\includegraphics[width=\textwidth]{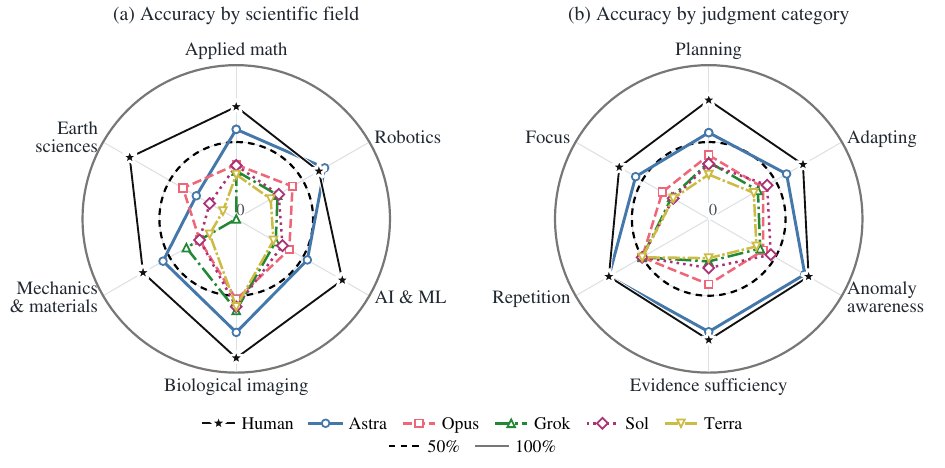}
\captionof{figure}{Accuracy by (a) scientific field (Table~\ref{tab:project-coverage}) and (b) judgment category (Table~\ref{tab:failure-taxonomy}).
Models run with the project record at medium reasoning effort.
Section~\ref{sec:human-reference} describes the human baseline.
Categories in (b) can overlap.}
\label{fig:radars}
\end{table*}

\subsection{What Should the Evaluation Preserve?}
We build cases retrospectively from completed scientific research projects.
Each project's senior researcher supervised it from the initial idea through the final write-up.
The full history links human interventions to subsequent investigation and provides evidence of their scientific value.
We use this evidence to recover useful corrections, then evaluate them at the earlier decision point.
Evaluators see only the earlier project record and two plans and judge which is more likely to advance the scientific question while the outcome is still unknown.
Providing the same plans isolates this judgment from generating plausible alternatives.
Figure~\ref{fig:construction} summarizes construction and evaluation.
\subsection{Recovering a Decision from Its History}
Candidate decisions come from human messages and changes to the project's human-instruction section.
The case writer inspects the full project history, recovers the original LLM plan, and selects a commit before the human intervention.
It reconstructs the human-corrected plan using the intervention and subsequent scientific investigation.
The evaluator receives the project record at that commit and the two plans, without the later history.
Accuracy measures how often the evaluator selects the reference plan.

Grok 4.6 wrote the cases.
Appendix~\ref{app:model-details} reports a controlled pilot in which the writer gains no advantage over the other models on its own cases.
The writing instructions exclude administrative actions, unrecoverable states or decisions, and human corrections that cannot be linked to an error.
Both options follow the same English \texttt{Goal:} and \texttt{Plan:} format.
The prompt requests comparable detail and length without wording that identifies the reference plan.
The evaluator does not see the writer's explanations.
Each case is linked to its commit and exact option text.

We selected 248 cases from 26 projects using 120 model responses per case collected during construction.
These responses cover five models, twelve conditions, and both option orders.
We retained cases whose model responses vary across the two plans.
Each retained case keeps a minimum share of responses on both plans.
Four one-sided cases are kept only after checking their reference plans against documentation at the case commit.
This selection focuses the comparison on cases where model responses vary.
Model-in-the-loop selection follows challenge-set precedent~\citep{kiela2021dynabench}.
The precedent applies to construction, and the resulting scores are diagnostic and measure recognition on decisions where models disagree rather than accuracy on research decisions in general.
Case review distinguishes evidence available at the decision point from later human corrections, outcomes, and changes in the senior researcher's goals.

\subsection{Recurring Failure Modes}
The human corrections reveal recurring failures across scientific research projects (Table~\ref{tab:failure-taxonomy}).
We name each category after the judgment involved and describe what goes wrong.
These judgments range from adapting to experimental results to deciding whether the evidence suffices for a conclusion.
The focus category covers experiments that can run correctly yet leave the research question unanswered.
For example, changing a cloth's physical area does not answer how its parameters should change with mesh resolution on the same cloth.
The coverage counts and radar scores group cases by these judgments.
All 248 cases have a primary judgment label, and some carry secondary labels.
Historical-error coding assigns 243 cases a primary error label and up to two secondary labels.
Coding uses the original LLM plans and human corrections, without the evaluated responses (Appendix~\ref{app:failure-cases}).

Construction requires a recoverable project state and documented human correction, so it applies wherever these records exist.
Table~\ref{tab:project-coverage} groups projects into six scientific fields by primary objective.
Biological research using neural models is therefore classified as a biological application.

\begin{table*}[t]
\centering
\small
\definecolor{comparisonwash}{RGB}{225,232,241}
\definecolor{comparisonink}{RGB}{0,68,136}
\setlength{\tabcolsep}{4.5pt}
\renewcommand{\arraystretch}{1.14}
\newlength{\comparisonwidth}
\setlength{\comparisonwidth}{\dimexpr\textwidth-6\tabcolsep\relax}
\begin{tabular}{@{}>{\raggedright\arraybackslash}p{.18\comparisonwidth}>{\raggedright\arraybackslash}p{.33\comparisonwidth}>{\raggedright\arraybackslash}p{.36\comparisonwidth}>{\raggedright\arraybackslash}p{.13\comparisonwidth}@{}}
\toprule
Benchmark & Scientific research task & Basis for evaluation & Fields \\
\midrule
PaperBench
& Replicate published ML experiments
& Replication rubrics from human authors
& ML \\[4pt]
RE-Bench
& Improve a defined ML R\&D objective
& Task scores normalized against human reference solutions
& ML \\[4pt]
MLE-bench
& Build ML solutions for Kaggle competitions
& Competition metrics and human leaderboards
& ML \\[4pt]
ScienceAgent\allowbreak{}Bench
& Write programs for scientific data workflows
& Expert-validated tasks; execution checks and rubrics
& Multiple \\[4pt]
SDE
& Reason about scientific scenarios; pursue discovery hypotheses
& Answers checked by human experts; computational oracles
& Multiple \\[4pt]
Corral
& Conduct scientific research; inspect tool traces
& Task scores and human expert diagnoses
& Multiple \\[4pt]
TruthInsight\allowbreak{}Bench
& Develop scientific claims from frozen scientific data
& Evidence rubric; bonus for source findings
& Multiple \\[4pt]
BLADE
& Choose and execute data analyses
& Analysis decisions checked by human experts
& Multiple \\[4pt]
ForeSci
& Choose scientific research directions using cutoff literature
& Later literature and targets checked by human experts
& ML \\
\midrule
\rowcolor{comparisonwash}
\textcolor{comparisonink}{\textbf{\bench{}}}
& Scientific planning and decision-making in real frontier scientific research
& \textbf{Senior researchers' corrections} and experimental evidence
& Multiple \\[4pt]
\bottomrule
\end{tabular}
\par\smallskip
{\color{comparisonink}\small
\textbf{\bench:}
Frontier scientific research $\rightarrow$ experimental evidence $\rightarrow$ human corrections $\rightarrow$ PI Judgment tests\par}
\caption{Representative scientific research benchmarks and how they establish good answers.
\bench tests scientific planning and decision-making using experimental records and senior researchers' corrections from real frontier scientific research.
Some benchmarks evaluate the listed activities in separate tasks.
Multiple means coverage across broad scientific fields.
AI/ML subfields count as one field.}
\label{tab:related}
\end{table*}

\subsection{Relation to Existing Benchmarks}
\label{sec:related}

Existing benchmarks assess different aspects of scientific research (Table~\ref{tab:related}).
PaperBench tests replication, while RE-Bench and MLE-bench test progress against defined research or engineering objectives~\citep{starace2025paperbench,wijk2025rebench,chan2025mlebench}.
ScienceAgentBench tests executable scientific workflows~\citep{chen2025scienceagentbench}.
The SDE benchmark includes scenarios from ongoing and past scientific research~\citep{song2026sde}; Corral supports scientific environments, trajectory analysis, and replay~\citep{riosgarcia2026corral}.
BLADE uses analysis decisions from human experts~\citep{gu-etal-2024-blade}; TruthInsightBench assesses evidence for scientific claims~\citep{yang2026truthinsightbench}; SoundnessBench uses human reviewers' judgments of proposals~\citep{ho2026soundnessbench}; and ForeSci evaluates plans against later literature~\citep{tian2026foresci}.

\bench evaluates whether models identify the better-supported scientific decision using information available before a human intervention.
Its reference plans are grounded in human corrections and the experimental records that explain them.
The AI Scientist systems and Agon develop workflows for conducting experiments and producing scientific manuscripts~\citep{lu2024aiscientist,yamada2025aiscientistv2,sun2026agon}.
\bench tests the scientific judgments needed to guide such workflows.

% One sentence per source line. Terminology, style rules, and figure placement: see CLAUDE.md.
\section{Can Models Recognize Human Corrections?}
\label{sec:results}
% GENERATED by scripts/build_paper_assets.py. Mirror any text change in that script, or a regeneration overwrites it. See CLAUDE.md.
% Generated by scripts/build_paper_assets.py; edit the generator.
\begin{table*}[t]
\centering
\small
\setlength{\tabcolsep}{5pt}
\begin{tabular}{@{}lcccl@{}}
\toprule
Model & Project (\%) & Options only (\%) & Best observed project (\%) & Setting \\
\midrule
Astra & 59.5 & 63.7 & 63.5 & xhigh \\
Opus & 38.9 & 58.9 & 46.8 & max \\
Grok & 33.1 & 54.0 & 38.1 & high \\
Sol & 35.1 & 46.8 & 37.7 & medium, after review 2 \\
Terra & 29.2 & 55.8 & 52.0 & medium, after review 1 \\
\midrule
Human baseline & \textbf{73.4} & --- & --- & agent assistant available \\
\bottomrule
\end{tabular}
\caption{Model accuracy on the same 248 cases, averaged over both option orders.
Project and Options only use medium effort.
Best observed project is the highest score among seven conditions.
These conditions are the five initial effort settings and the first and second review at medium effort.
Complete 95\% intervals are in Appendix~\ref{app:model-details}.
Appendix~\ref{app:human-reference} describes the human baseline.
The options-only condition measures model context sensitivity.}
\label{tab:main-results}
\end{table*}

\subsection{Evaluation Setup}
\label{sec:protocol}
We evaluate GPT-6 Astra~\citep{openai2026gpt6astra}, Claude Opus 5~\citep{anthropic2026opus5}, Grok 4.6~\citep{xai2026grok46}, and GPT-5.6 Sol and Terra~\citep{openai2026gpt56}, abbreviated as Astra, Opus, Grok, Sol, and Terra.
In the \emph{project} condition, the model receives two plans and the project record.
We ask it to start with the research idea, proposal, and progress notes, then consult other files as needed.
The project record excludes Git history and later project state.
Network access and research execution are prohibited.
The model returns a choice and an explanation.
The \emph{options-only} condition presents the same plans without the project record.
Client details appear in Appendix~\ref{app:model-details}.
Appendix~\ref{app:evaluation-prompts} gives the exact prompts, all twelve conditions, and available materials.

The main comparison requests medium reasoning effort and evaluates every case in two separate sessions with reversed option order.
We average correctness over the two orders, then give each case equal weight.
Chance accuracy is 50\%.
Completed responses without a parseable choice count as incorrect, although all responses in the main comparison have a parsed choice.
We estimate 95\% intervals from 10,000 paired bootstrap samples of cases and report project-cluster intervals in the appendix.

\paragraph{Human baseline.}
\label{sec:human-reference}
The human baseline comes from fresh researchers with no prior exposure to these projects and no involvement in supervising them.
They answered through an interface for reading the project, with an agent assistant available.
Each learned the project from the project record and judged a case, making the final choice.
Thirteen PhD students and postdocs took part, and all were informed about the study, gave consent, and are reported under anonymous identifiers.
Appendix~\ref{app:human-reference} describes the human baseline protocol.

\subsection{Models Often Reject the Human-Corrected Plan}
\label{sec:main-results}
At medium effort with the project record, four of the five models select the rejected plan more often than the reference plan on this selected set (Table~\ref{tab:main-results}).
For all four, the upper bounds of both case and project bootstrap intervals are below 50\%.
Astra reaches 59.5\%, and the five-model average is 39.2\%.
The highest observed project score across all tested effort settings is Astra's 63.5\% at xhigh, compared with 73.4\% for the human baseline.

On 49 cases, all five models choose the rejected plan in both orders; on 16, they all choose the reference plan.
The original LLM plan is on-policy text, while the human-corrected plan is off-policy.
A systematic preference for the original plan is the structural signature of this asymmetry, not a coincidence.

\begin{figure}[t]
\centering
\includegraphics[width=\columnwidth]{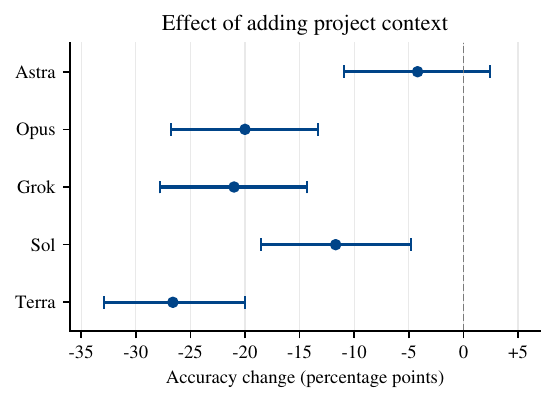}
\caption{Accuracy difference with the project record, relative to the options-only condition.
Comparisons pair the same cases at medium effort.
Bars show 95\% case-bootstrap intervals.}
\label{fig:context-effect}
\end{figure}

\begin{figure*}[t]
\centering
\includegraphics[width=\textwidth]{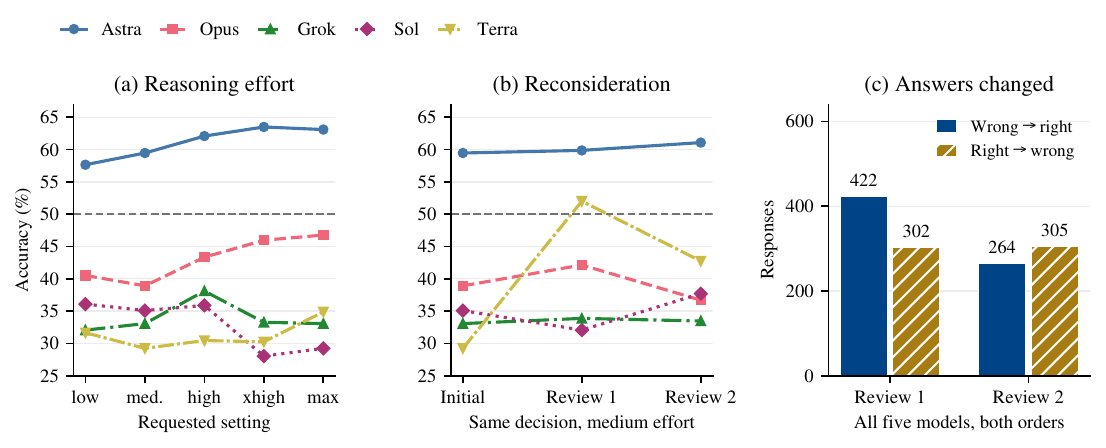}
\caption{More deliberation can restore a rejected plan after a correct choice.
(a) Accuracy at each requested effort setting.
(b) The initial answer at medium effort and its two recorded continuations.
(c) Repairs and new errors across five models and both orders.
Each review contains 2,480 response transitions on 248 cases.
Panels (a) and (b) use a 25--67\% accuracy range.
The dashed line marks 50\%.
Effort settings are not matched token budgets.
Repeated responses are not independent cases.}
\label{fig:reasoning-settings}
\end{figure*}

\subsection{Different Judgments Reveal Different Gaps}
\label{sec:failure-results}
Figure~\ref{fig:radars} breaks down the variation across types of scientific decisions by scientific field (a) and by the judgment each case requires (b).
Each model point averages the two orders within a case and weights cases equally within its group.
Astra reaches 72.2\% on anomaly-awareness cases, but 56.1\% on planning cases and 54.9\% on focus cases.

Two cases illustrate the judgment needed to choose an experimental scale (Appendix~\ref{app:failure-cases}).
In case 21, an untested setup calls for a pilot, yet models retain a 200-item-per-cell batch in seven of ten responses.
In case 13, testing the hypothesis requires a full-length training run, yet models retain three short gated runs in six of ten responses.
Models must allocate effort according to what the investigation requires, rather than consistently prefer larger or smaller experiments.

\subsection{Lower Accuracy with the Project Record}
\label{sec:context-results}
The project record contains observations alongside the original agent's explanations, including mistaken interpretations.
On this selected set, every model nevertheless scores higher without the project record at medium effort.
Mean accuracy is 39.2\% with the project record and 55.8\% without it (Figure~\ref{fig:context-effect}).
For four models, the paired 95\% interval of this decrease excludes zero, and Astra's includes zero.
Long contexts can lower task accuracy even when the added material is informative~\citep{liu2023lost}.
The comparison evaluates the complete protocol of the project condition, including retrieval, interpretation, and the client-specific reading interface.

The Transformer case illustrates how an inherited explanation can make the rejected plan appear justified.
Opus accepts the progress note's claim that scoring faults were excluded and uses the record to support the original mistake by treating that claim as established evidence.
PI Judgment requires checking the claim against the observations and code.

\subsection{More Thinking Can Lead Back to the Same Mistake}
\label{sec:reasoning-results}
Test-time scaling can improve model performance by allocating more computation to reasoning~\citep{snell2024scaling}.
Can further deliberation help models question these explanations and retain a correction?
We vary requested effort from low through max and continue the original medium-effort session twice.
Each continuation permits further project reading and a changed answer, without correctness feedback.
Refinement methods that supply external feedback can improve outputs~\citep{madaan2023self,shinn2023reflexion}, while reconsideration here receives no feedback, and self-correction without external feedback can reduce accuracy~\citep{huang2023large}.
We pair each answer with its recorded parent to count both repaired and newly introduced errors.

Higher effort sometimes helps, but the gains are inconsistent (Figure~\ref{fig:reasoning-settings}a).
From low to max, four models improve while Sol declines from 36.1\% to 29.2\%.
Terra rises from 29.2\% to 52.0\% after the first review, then falls to 42.7\% after the second.
Opus also loses accuracy on the second review, and both declines exclude zero under paired case and project intervals.

Across the five models, the first review repairs 422 errors and introduces 302.
The second repairs 264 and introduces 305, partially undoing the earlier gain.
Mean accuracy across the three stages is 39.2\%, 44.0\%, and 42.3\%, with paired intervals in Table~\ref{tab:reasoning-transitions}.

In the Transformer case, Opus first chooses the larger model, then questions the scorer and chooses the reference plan.
On its second review, it returns to the larger model with a premise that the frozen code contradicts.

% One sentence per source line. Terminology, style rules, and figure placement: see CLAUDE.md.
\Needspace{16\baselineskip}
\section{From Diagnosis to Better AutoResearch Agents}
\label{sec:conclusion}

Sustained scientific research exposed decisions that frontier models and agentic systems still left to senior researchers.
\bench turns these human corrections into tests of PI Judgment, such as adapting to experimental results and keeping experiments on the research question.
Even with the project record and further deliberation, models often return to the decisions senior researchers corrected.

Each case supplies a project record, a preferred human-corrected plan, and a rejected original LLM plan.
These comparisons match the training format of direct preference optimization~\citep{rafailov2023dpo}.
As future work, a preference scorer trained on these comparisons could supply rewards for research actions in agentic reinforcement learning.
The next test is whether such training improves decisions in new projects and reduces human intervention.

Learning these shared judgments could support more autonomous scientific AutoResearch agents across disciplines, including research aimed at improving AI systems themselves.

% One sentence per source line. Terminology, style rules, and figure placement: see CLAUDE.md.
\section*{Limitations}

\bench covers theoretical and computational research using CPUs and GPUs, including simulation and analysis of existing data such as biological images.
Wet-lab execution and workflows using specialized commercial scientific software such as COMSOL remain outside this evaluation.
The cases test judgments central to PhD research across disciplines, including experiment selection, anomaly diagnosis, and revision of conclusions.
Evaluation in other settings would test how these judgments transfer alongside different tools and experimental procedures.

\bench measures recognition of corrective scientific decisions drawn from projects supervised by senior researchers and selected using the procedure in Section~\ref{sec:construction}.
Generating plans and carrying out scientific research require separate evaluation.
The selected cases do not estimate how often these failures occur across all scientific research activity.
Several cases come from each project, so we report project-cluster intervals alongside case-level intervals.

The evaluated clients expose different reading tools, and requested effort settings do not specify equal token budgets.
Our effort and reconsideration experiments test these available controls; they do not test controlled compute scaling or verifier-guided search.

% One sentence per source line. Terminology, style rules, and figure placement: see CLAUDE.md.
\section*{Ethics Statement}

The source records include scientific research conversations and project histories.
Public materials should preserve the evidence needed to understand each decision while removing personal identifiers, account information, and unrelated private discussion.
All fresh researchers were informed about the study, gave consent, and are reported under anonymous identifiers; Appendix~\ref{app:human-reference} describes the evaluation setting.

Language models assisted with drafting, organizing the literature, and writing analysis code.
The authors remain responsible for checking the claims, references, and reported results.

\paragraph{Data availability.}
The complete question set of 248 cases with anonymized case identifiers, the two options per case in their exact wording, and reference answers is released in the \bench repository~\citep{scipivot2026}.
The repository also hosts the exact evaluation and case-writing prompts and aggregate model results for all twelve conditions.
Individual model response logs are not part of this release.
The frozen project records are too large to upload, and the same repository will host them.
All case details will be released under an open license that permits research and training use, in JSON and plain-text formats.

\bibliography{references}
\clearpage
\appendix
% GENERATED by scripts/build_paper_assets.py. Mirror any text change in that script, or a regeneration overwrites it. See CLAUDE.md.
% One sentence per source line. Terminology, style rules, and figure placement: see CLAUDE.md.
% Generated by scripts/build_paper_assets.py; one sentence per source line.
\section{Model Results and Statistical Details}
\label{app:model-details}
\subsection{Project-Reading Clients}
The requested model identifiers are \texttt{gpt-6-astra}, \texttt{claude-opus-5}, \texttt{grok-4.6}, \texttt{gpt-5.6-sol}, and \texttt{gpt-5.6-terra}.
Opus and Grok use dedicated file-reading and search tools through one shared client harness, the Claude client.
Shell execution, editing, and web tools are disabled in that harness.
The GPT models use shell commands through Codex, with a read-only project and web search disabled.
All project prompts prohibit network access and reading outside the project record.
The options-only prompt removes the project record and its reading instructions while keeping the same plans.
The comparison between conditions therefore includes retrieval, interpretation, and the client-specific tool interface.

\subsection{Scoring and Uncertainty}
The benchmark contains 248 cases from 26 projects, with two option orders per case.
Each medium-effort Project or Options only cell therefore contains 496 completed responses.
We retain the last logged record for each combination of case, option hash, condition, engine, requested model, effort, prompt hash, and repeat identifier.
This rule replaces a failed or superseded attempt rather than counting a retry as another observation.
All 29,760 retained runs completed without a transport error.
Eight responses could not be parsed and receive a score of zero.
These are four Sol responses at xhigh, one Sol response at max, one Terra response at high, one Grok response at xhigh, and one Grok response after the second review.
All eight unparsed responses occur in project-condition runs.
Both medium-effort primary conditions are fully parsed.
Excluding unparsed responses changes any affected cell by at most 0.23 percentage points.

For each case, we average the two binary scores before estimating uncertainty.
The main intervals use 10,000 percentile bootstrap samples of 248 cases drawn with replacement.
The random seed is 20261004.
We resample the same case indices in both conditions when estimating their difference.
The two option orders are repeated measurements, not independent cases.
Table~\ref{tab:medium-confidence-intervals} reports the accuracy of both conditions and their paired difference, with complete case-bootstrap intervals.
\begin{table*}[t]
\centering
\small
\setlength{\tabcolsep}{8pt}
\begin{tabular}{@{}lccc@{}}
\toprule
Model & Project (\%) & Options only (\%) & Paired difference (pp) \\
\midrule
Astra & 59.5 [53.8, 65.1] & 63.7 [58.5, 69.0] & -4.2 [-10.9, +2.4] \\
Opus & 38.9 [33.5, 44.4] & 58.9 [54.0, 63.7] & -20.0 [-26.8, -13.3] \\
Grok & 33.1 [28.0, 38.1] & 54.0 [48.6, 59.5] & -21.0 [-27.8, -14.3] \\
Sol & 35.1 [29.8, 40.3] & 46.8 [41.3, 52.0] & -11.7 [-18.5, -4.8] \\
Terra & 29.2 [24.6, 34.1] & 55.8 [50.6, 60.9] & -26.6 [-32.9, -20.0] \\
\bottomrule
\end{tabular}
\caption{Medium-effort accuracies and paired Project minus Options only differences on 248 cases.
Brackets give 95\% percentile case-bootstrap confidence intervals after averaging the two option orders.}
\label{tab:medium-confidence-intervals}
\end{table*}

Several cases come from the same project and may describe related human interventions.
We therefore sample 26 projects with replacement in each bootstrap replicate and include every case from each sampled project.
A project selected more than once contributes its complete case set each time.
No additional sampling occurs within projects.
We continue to weight cases equally and use the same project draws in both conditions.
Table~\ref{tab:project-sensitivity} reports project-cluster intervals for project accuracy and the difference between conditions.
\begin{table}[t]
\centering
\small
\setlength{\tabcolsep}{3pt}
\begin{tabular}{@{}lcc@{}}
\toprule
Model & Project 95\% CI & Difference 95\% CI \\
\midrule
Astra & [52.6, 67.0] & [-11.8, +3.6] \\
Opus & [33.4, 45.2] & [-25.0, -14.5] \\
Grok & [28.3, 39.2] & [-25.6, -15.7] \\
Sol & [29.5, 42.6] & [-17.8, -4.5] \\
Terra & [24.2, 36.8] & [-32.6, -19.0] \\
\bottomrule
\end{tabular}
\caption{Project-cluster 95\% intervals for project accuracy (\%) and the Project minus Options only difference (percentage points).
Both estimates retain equal case weights.}
\label{tab:project-sensitivity}
\end{table}

\subsection{Project Coverage and Effort Settings}
Tables~\ref{tab:project-sizes} and~\ref{tab:project-sizes-continued} list each project's identifier, research question, area, and case count.
We classify projects by their primary scientific objectives.
The construction procedure does not depend on these area labels.
Project size ranges from one to 43 cases, so averaging project means differs from averaging cases.
The effort labels in Table~\ref{tab:all-efforts} are the requested effort settings that the evaluated model interfaces expose.
We do not interpret them as equal token budgets or equal compute across providers.
Every entry uses the same 248 cases and two orders per case.
PDE abbreviates partial differential equation.
\begin{table*}[t]
\centering
\footnotesize
\setlength{\tabcolsep}{3pt}
\renewcommand{\arraystretch}{1.12}
\begin{tabularx}{\textwidth}{@{}>{\raggedright\arraybackslash}p{.24\textwidth}>{\raggedright\arraybackslash}X>{\raggedright\arraybackslash}p{.15\textwidth}r@{}}
\toprule
Project identifier & Scientific question & Area & Cases \\
\midrule
\texttt{cpd-\allowbreak{}lattice} & Can screened low-rank function proposals accelerate high-dimensional approximation and PDE solving? & Applied mathematics \& scientific computing & 4 \\
\texttt{fex-\allowbreak{}dim-\allowbreak{}lift-\allowbreak{}skeleton} & When can solution structures found for low-dimensional PDEs be reused in higher dimensions? & Applied mathematics \& scientific computing & 12 \\
\texttt{fex-\allowbreak{}foundation-\allowbreak{}model} & When does a pretrained expression-search prior help solve a new PDE? & Applied mathematics \& scientific computing & 12 \\
\texttt{fex-\allowbreak{}search-\allowbreak{}complexity} & When can a search algorithm efficiently discover a PDE solution that has a compact expression? & Applied mathematics \& scientific computing & 4 \\
\texttt{fex-\allowbreak{}synchro-\allowbreak{}prior} & Can checked spectral priors resolve frequency-search bottlenecks in symbolic PDE solvers? & Applied mathematics \& scientific computing & 1 \\
\texttt{ocl-\allowbreak{}calibration} & How can calibration ensure that optimizer-selected decisions satisfy learned constraints? & Applied mathematics \& scientific computing & 8 \\
\texttt{sequential-\allowbreak{}pinn} & Can staged training make deep physics-informed networks more reliable and efficient PDE solvers? & Applied mathematics \& scientific computing & 43 \\
\texttt{sr-\allowbreak{}main} & How reliably do symbolic regression methods recover equations under a shared evaluation protocol? & Applied mathematics \& scientific computing & 1 \\
\texttt{contact-\allowbreak{}error-\allowbreak{}budget} & Which errors in reconstructed reference motions predict failed humanoid tracking? & Robotics \& control & 8 \\
\texttt{rule-\allowbreak{}system-\allowbreak{}robot-\allowbreak{}control} & Can agents maintain explicit program controllers that reliably imitate humanoid motions? & Robotics \& control & 10 \\
\texttt{soft-\allowbreak{}support-\allowbreak{}bfm} & How can validated soft-support simulation help humanoids roll over and stand up on deformable surfaces? & Robotics \& control & 33 \\
\texttt{vln-\allowbreak{}verbal-\allowbreak{}route} & Can named places and text routes support reliable vision-and-language navigation? & Robotics \& control & 7 \\
\bottomrule
\end{tabularx}
\caption{Benchmark projects (1/2).
The questions describe scientific objectives rather than established findings.
Area labels describe the current sample.
We do not count secondary methodological links separately.}
\label{tab:project-sizes}
\end{table*}
\begin{table*}[t]
\centering
\footnotesize
\setlength{\tabcolsep}{3pt}
\renewcommand{\arraystretch}{1.12}
\begin{tabularx}{\textwidth}{@{}>{\raggedright\arraybackslash}p{.24\textwidth}>{\raggedright\arraybackslash}X>{\raggedright\arraybackslash}p{.15\textwidth}r@{}}
\toprule
Project identifier & Scientific question & Area & Cases \\
\midrule
\texttt{cd-\allowbreak{}projection-\allowbreak{}ladder} & Does a failed reasoning trace disrupt later answers through problem-specific errors or general context interference? & AI \& machine learning & 1 \\
\texttt{cot-\allowbreak{}mon-\allowbreak{}stego-\allowbreak{}canary} & How much does reading reasoning text improve misconduct detection beyond observing final actions? & AI \& machine learning & 4 \\
\texttt{elf-\allowbreak{}ecf} & Can flow matching improve sentence-level language generation with fewer sampling steps? & AI \& machine learning & 12 \\
\texttt{elf-\allowbreak{}fk-\allowbreak{}steer} & Can intermediate text previews guide generation toward factual and otherwise constrained outputs? & AI \& machine learning & 3 \\
\texttt{hybrid-\allowbreak{}scaling-\allowbreak{}decomp} & Do model-selection criteria systematically favor particular neural scaling laws when the data cannot distinguish them? & AI \& machine learning & 4 \\
\texttt{looped-\allowbreak{}lora} & Can small trainable adapters turn a frozen pretrained Transformer into an effective recurrent model? & AI \& machine learning & 24 \\
\texttt{muon-\allowbreak{}mp-\allowbreak{}spectrum} & Can random-matrix spectra explain neural gradients and improve Muon optimizer settings? & AI \& machine learning & 1 \\
\texttt{prc-\allowbreak{}real-\allowbreak{}prompt-\allowbreak{}delete-\allowbreak{}helps} & How often does deleting a sentence from a natural prompt improve task performance? & AI \& machine learning & 4 \\
\texttt{transformeron} & Does critical behavior in generated text also appear in internal Transformer measurements? & AI \& machine learning & 6 \\
\texttt{kaggle-\allowbreak{}biohub} & How can 3D time-lapse microscopy reveal cell trajectories and divisions to reconstruct lineages? & Biological imaging & 21 \\
\texttt{cloth-\allowbreak{}renormalization} & How should cloth simulation parameters change across resolutions to preserve realistic motion? & Mechanics \& materials & 9 \\
\texttt{sr-\allowbreak{}material} & Which search methods discover useful descriptors of crystal energy differences and metallic behavior? & Mechanics \& materials & 11 \\
\texttt{earth-\allowbreak{}back-\allowbreak{}proj} & How much do physical consistency corrections and residual learning each improve precipitation downscaling? & Earth sciences & 1 \\
\texttt{kaggle-\allowbreak{}rogii} & How can drilling observations and reference wells predict geology along a horizontal wellbore? & Earth sciences & 4 \\
\bottomrule
\end{tabularx}
\caption{Benchmark projects (2/2).
The questions describe scientific objectives rather than established findings.
Area labels describe the current sample.
We do not count secondary methodological links separately.}
\label{tab:project-sizes-continued}
\end{table*}
\begin{table*}[t]
\centering
\small
\begin{tabular}{@{}lccccc@{}}
\toprule
Model & Low & Medium & High & xhigh & Max \\
\midrule
Astra & 57.7 / 64.5 & 59.5 / 63.7 & 62.1 / 64.9 & 63.5 / 64.5 & 63.1 / 62.7 \\
Opus & 40.5 / 59.5 & 38.9 / 58.9 & 43.3 / 59.1 & 46.0 / 58.9 & 46.8 / 57.9 \\
Grok & 32.1 / 53.8 & 33.1 / 54.0 & 38.1 / 55.8 & 33.3 / 56.5 & 33.1 / 53.4 \\
Sol & 36.1 / 54.4 & 35.1 / 46.8 & 35.9 / 54.2 & 28.0 / 53.8 & 29.2 / 55.8 \\
Terra & 31.7 / 56.5 & 29.2 / 55.8 & 30.4 / 55.4 & 30.2 / 55.6 & 34.9 / 57.3 \\
\bottomrule
\end{tabular}
\caption{Accuracy (\%) at every requested effort setting.
Each cell reports Project / Options only.
Unparsed completed responses count as incorrect.}
\label{tab:all-efforts}
\end{table*}

\subsection{Option Order and Reconsideration}
The two orders place the reference plan in position A and position B, respectively.
Table~\ref{tab:option-order} reports each order separately at medium effort.
The main score gives the two orders equal weight instead of selecting the better order.
The A/B differences can reflect a position preference or an interaction between option order and reasoning.

The reconsideration experiment follows the project-condition answer through two continuation stages at medium effort.
We join each child to the logged parent run and attempt, then verify that parent and child runs agree on the case, option hash, commit, model, engine, effort, repeat, and order.
Each model has 496 parent-child pairs at each stage.
The change at a stage is the number of corrected errors minus the number of newly introduced errors, divided by 496.
The accuracy trajectories in Figure~\ref{fig:reasoning-settings} use these actual chains rather than independently regenerated initial answers.
We count changes in the selected plan rather than changes in the underlying scientific reasoning.
Table~\ref{tab:reasoning-transitions} reports paired changes after the second review, with uncertainty estimated from 100,000 resamples.
For the five-model mean, we average the fixed models within each case before resampling cases or whole projects.
% Computed on the 248 benchmark cases by scripts/recompute_reasoning_transitions.py;
% validated against the earlier full-set audit table before rescoping.
\begin{table*}[t]
\centering
\small
\setlength{\tabcolsep}{6pt}
\begin{tabular}{@{}lrcc@{}}
\toprule
Model & $\Delta$ (pp) & Paired case 95\% CI & Project-cluster 95\% CI \\
\midrule
Astra & +1.2 & [-0.8, +3.2] & [-0.9, +2.9] \\
Opus & -5.4 & [-8.7, -2.2] & [-9.6, -1.9] \\
Grok & -0.4 & [-4.0, +3.2] & [-3.0, +3.0] \\
Sol & +5.6 & [+0.2, +11.1] & [+1.2, +10.0] \\
Terra & -9.3 & [-15.3, -3.0] & [-13.4, -4.3] \\
\midrule
Five-model mean & -1.7 & [-3.5, +0.2] & [-2.7, -0.3] \\
\bottomrule
\end{tabular}
\caption{Accuracy change after the second review, relative to the first, in percentage points.
We average both option orders within each of 248 cases before computing paired differences.
Intervals use 100,000 paired percentile bootstrap samples.
The project bootstrap resamples 26 whole projects with replacement and retains case weights.
The mean averages the fixed set of five evaluated models within each case.
}
\label{tab:reasoning-transitions}
\end{table*}

\subsection{Writer Sensitivity}
One writer model wrote all released cases under one instruction set, so writer identity is constant across every model in the main comparison.
A controlled pilot checks whether this choice favors the writer model.
The pilot uses three writers, Astra, Opus, and Grok.
Each writer produced options for 20 shared cases, with two option orders and three answering models.
Seven shared cases have project records frozen at different commits across writers.
We therefore analyze the 13 cases whose frozen commits and option hashes match across all three writers (Table~\ref{tab:writer-same-commit}).
Every cell contains 26 responses.
\begin{table*}[!tp]
\centering
\begin{minipage}[t]{.48\textwidth}
\centering
\small
\setlength{\tabcolsep}{3pt}
\begin{tabular}{@{}lcc@{}}
\toprule
Model & Project (A / B) & Options only (A / B) \\
\midrule
Astra & 54.8 / 64.1 & 56.5 / 71.0 \\
Opus & 37.1 / 40.7 & 44.0 / 73.8 \\
Grok & 41.9 / 24.2 & 61.7 / 46.4 \\
Sol & 38.7 / 31.5 & 51.2 / 42.3 \\
Terra & 32.7 / 25.8 & 54.4 / 57.3 \\
\bottomrule
\end{tabular}
\caption{Accuracy (\%) when the reference plan appears first (A) or second (B), with 248 responses per order.}
\label{tab:option-order}
\end{minipage}
\hfill
\begin{minipage}[t]{.48\textwidth}
\centering
\small
\setlength{\tabcolsep}{4pt}
\begin{tabular}{@{}llccc@{}}
\toprule
Writer & Context & Astra & Opus & Grok \\
\midrule
Astra & Project & 42.3 & 19.2 & 11.5 \\
Astra & Options & 53.8 & 38.5 & 30.8 \\
Opus & Project & 53.8 & 42.3 & 34.6 \\
Opus & Options & 76.9 & 65.4 & 50.0 \\
Grok & Project & 50.0 & 34.6 & 23.1 \\
Grok & Options & 61.5 & 69.2 & 61.5 \\
\bottomrule
\end{tabular}
\caption{Controlled writer pilot on 13 cases whose project records are frozen at the same commit for all three writers, with 26 responses per cell.
Columns identify the answering model, and entries are accuracy (\%).
Options denotes the options-only condition.}
\label{tab:writer-same-commit}
\end{minipage}
\end{table*}

In the pilot, the writer of the released set has no advantage over its peers.
Grok, the writer of the released set, is never the top scorer on its own cases.
Opus reaches 34.6\% on Grok-written cases with the project record, against 23.1\% for Grok itself, and Opus also leads the options-only cells.
The largest single gain runs across writers rather than along them.
Astra reaches its highest accuracy, 76.9\%, on Opus-written cases.
Writer identity changes absolute case difficulty, but the writer of the released set never outscores its peers on its own text.
The main comparison holds the writer and option texts fixed across all answering models, so writer identity does not confound the cross-model comparison.

% One sentence per source line. Terminology, style rules, and figure placement: see CLAUDE.md.
% The Goal/Plan blocks below preserve the released option wording; only LaTeX escaping and sentence line breaks are added.
\section{Two Illustrative Scientific Decisions}
\label{app:showcases}

These examples show the decisions that human interventions changed and quote evaluation responses that repeated the original mistake.
Table~\ref{tab:showcase-results} reports the reference-plan choices of all five models in both option orders for each example.
The reference and rejected plans below reproduce the benchmark task wording.

\subsection{Checking Evaluation Before Changing Model Size}
\label{app:showcase-training}

The project studied whether repeated passes through a language-model block improve performance.
It was building an evaluation set with several difficulty levels for the selected 0.8B base model.
At the frozen decision, three LiveCodeBench groups~\citep{jain2024livecodebench} scored 0/480.
A project note treated earlier checks as having ruled out an evaluation fault and proposed rerunning the tasks on a 2B model.
The frozen code, however, passed raw generated text directly to the new tasks' graders without the official answer postprocessing.
Its preflight also did not reject individual benchmark groups with all-zero scores.
The senior researcher chose to retain 0.8B and find suitable official tasks.
A subsequent audit identified the scoring fault, which the reference plan addressed before further calibration.
Later regrading used the same 608 affected completions, verified byte for byte, and recovered 70 correct answers among the 480 code items.
The same four difficulty cells, however, remained unfilled.
In one response under the project condition, Opus nevertheless selected the larger-model route and wrote, ``The blocker is not an evaluation bug, it is a size/tier conflict.''
Holding tasks fixed was a reasonable control in principle.
But accepting the unverified claim that scoring faults had been excluded made the original zero scores an unreliable basis for the size decision.

For this option order, two subsequent reconsiderations moved the choice from the rejected plan to the reference plan and back to the rejected plan.
The project record and option order were unchanged throughout.
The first reconsideration acknowledged, ``I previously trusted STATE.md's assertion that the evaluator was ruled out,'' and selected the reference plan after examining the pattern of zero scores.
That selection was correct, but the explanation behind it was only partly correct.
It relied in part on a comparison of free-generation scores with forced-choice scores, rather than directly identifying the missing postprocessing in the code.
The second reconsideration noticed this difference in evaluation format.
It nevertheless returned to the rejected plan, claiming that the nonzero results on MBPP, the Mostly Basic Python Problems benchmark~\citep{austin2021mbpp}, used an ``identical generate/extract/execute pipeline.''
The frozen code contradicts that equivalence.
The old-pool path used for MBPP truncates generated text before grading, whereas the new-pool path used for LiveCodeBench passes raw text to its graders.
Success on the MBPP path therefore did not establish that the LiveCodeBench path was correct.
The sequence connects both answer changes to the evidence Opus cited.
Further review repaired the choice once, then reinstated the rejected plan using a faulty comparison.
The quoted plans name the following evaluation suites.
LiveCodeBench covers code generation~\citep{jain2024livecodebench}, MATH covers arithmetic~\citep{hendrycks2021math}, and CRUXEval covers code reasoning and execution~\citep{gu2024cruxeval}.
ASDiv covers arithmetic word problems~\citep{miao2021asdiv}, and SciBench covers scientific problem solving~\citep{wang2023scibench}.
GPQA abbreviates Graduate-Level Google-Proof QA~\citep{rein2023gpqa}.

\paragraph{Reference plan.}
\textbf{Goal:} Finish the real-task tier asset after the 0.8B Base-K1 calibration stopped at 8 of 12 cells.

\textbf{Plan:} Keep 0.8B as the selected size, and keep the 25, 50, and 75 percent accuracy layers, the quotas, and the official metrics.
Grade generations with the official LiveCodeBench stop and extract functions, and stop a panel when any benchmark group scores all zeros.
Regrade or regenerate the LiveCodeBench and word-sorting groups.
Add official CRUXEval, ASDiv, and SciBench sources on this 0.8B Base.
Write the 12-cell asset after all 12 cells are filled.
Start the four matched-depth micro-runs after that asset exists.

\paragraph{Rejected plan.}
\textbf{Goal:} Finish the real-task tier asset after the 0.8B Base-K1 calibration stopped at 8 of 12 cells.

\textbf{Plan:} Keep the current LiveCodeBench, MATH, and GPQA items.
Score the same registered code, math, and science subset on the 2B Base with the same calibration plan and the same method that graded the 0.8B outputs, using about 1700 rows.
If 2B fills the missing cells, set the selected experiment size to 2B and write the 12-cell asset.
If 2B still misses cells, widen the accuracy layers on 0.8B or change the official tasks after that measurement, then write the asset.
Start the four matched-depth micro-runs after that asset exists.

\subsection{Testing a Restrictive Physical Model}
\label{app:showcase-cloth}

The project aimed to calibrate a cloth simulator against recorded motion before studying how its fitted parameters change with simulation resolution.
Its frozen proposal specified six positive parameters and a nonzero reference vector, but the working plan fixed four parameters to zero and searched only the remaining two.
All ten sampled minima reported in the project record lay at the same corner of their search domains.
The senior researcher clarified that two reference values came from published fits and four from simulator defaults, so missing fitted material values did not justify setting the corresponding physical parameters to zero.
The reference plan therefore reopened the six-parameter fit while retaining the original two-parameter slice as a comparison on the same objective.
The defaults supplied a search center for testing the fuller physical model.
Opus described the restricted search as a ``standing, human-authorized restriction'' and selected it in both option orders under the project condition.
In one response, it also stated, ``Option B is the direct continuation of the actual live plan.''
Its concerns about previous optimization instability and cost were relevant, but the continuing boundary results called for testing the restriction itself.
RG abbreviates renormalization group.

\paragraph{Reference plan.}
\textbf{Goal:} Identify the six simulator parameters on one time-step cell, using the six-parameter reference vector as the search center.

\textbf{Plan:} Keep density at 0.3046 and search delta, alpha, kappa, kappa\_bnd, stretch, and shear together.
Center the search on phi\_ref = (0.3046, 0.6092, 1.523e-4, 1.523e-5, 1e-6, 1e-3).
Use the four full marker recordings as one joint Eq. (10) objective on the native G0 mesh with eight substeps.
Alternate two-dimensional blocks in the order dynamic, bending, then in-plane, run that cycle twice, and then run a bounded joint Nelder-Mead polish.
Fit a second arm that searches only alpha and delta with the other four at zero, on the same objective.
Start cross-scale RG work when the six-parameter endpoints sit off the box walls and agree across starts.

\paragraph{Rejected plan.}
\textbf{Goal:} Map how the joint two-parameter loss surface for alpha and delta changes across allowed time steps, with the other four physical parameters held at zero.

\textbf{Plan:} Keep density at 0.3046 and set kappa, kappa\_bnd, stretch, and shear to zero.
Use the four full marker recordings as one joint Eq. (10) objective on the native G0 mesh.
After reading the five-scale coarse surfaces and the external corner bridge, add the remaining allowed time-step levels and extra shifted grids on the same two-parameter box of alpha and delta.
Report sampled minima, edge profiles, and the direction of the surface at the walls.
Start a lowest-order RG fit when at least five time-step levels have interior minima in this two-parameter slice.

\begin{table}[t]
\centering
\small
\setlength{\tabcolsep}{4pt}
\begin{tabular}{lcccc}
\toprule
& \multicolumn{2}{c}{Language model} & \multicolumn{2}{c}{Cloth simulator} \\
\cmidrule(lr){2-3}\cmidrule(lr){4-5}
Model & Project & Options & Project & Options \\
\midrule
Astra & 2/2 & 1/2 & 0/2 & 2/2 \\
Opus & 1/2 & 2/2 & 0/2 & 0/2 \\
Grok & 2/2 & 2/2 & 1/2 & 0/2 \\
Sol & 2/2 & 1/2 & 0/2 & 0/2 \\
Terra & 0/2 & 1/2 & 0/2 & 1/2 \\
\bottomrule
\end{tabular}
\caption{Reference-plan choices across the two option orders at medium effort.
``Options'' denotes the options-only condition.
Each model answers the same case twice, with the option order reversed.
The ten choices per condition are not ten independent cases.}
\label{tab:showcase-results}
\end{table}

% One sentence per source line. Terminology, style rules, and figure placement: see CLAUDE.md.
\section{Judgment Categories and Illustrative Cases}
\label{app:failure-cases}

% GENERATED by scripts/build_paper_assets.py. Mirror any text change in that script, or a regeneration overwrites it. See CLAUDE.md.
% Generated by scripts/build_paper_assets.py from prepared medium-effort case scores.
\begin{table*}[t]
\centering
\small
\setlength{\tabcolsep}{6pt}
\begin{tabular}{@{}rlcc@{}}
\toprule
Case & Decision shorthand & Project & Options only \\
\midrule
206 & Check stiffness and travel limits & 10/10 & 7/10 \\
3 & Start the resolution matrix; drop the baseline gate & 8/10 & 3/10 \\
21 & Pilot the API pipeline at 20 items per cell & 3/10 & 6/10 \\
13 & Train one seed for the full 8,000 iterations & 4/10 & 6/10 \\
\bottomrule
\end{tabular}
\caption{Reference-plan choices at medium effort for the four examples in Appendix~\ref{app:failure-cases}.
We selected these examples for their decision structure before inspecting model scores.
Each count is out of ten responses from five fixed models and two option orders, not ten independent cases.
Case numbers are zero-based indices in the benchmark.}
\label{tab:case-observations}
\end{table*}

\subsection{Coding the Historical Decision}
The codebook describes the original LLM plan and the decision that the human intervention changed, without using the evaluated models' answers.
We organize the codebook around three overlapping demands, namely identifying the research goal, interpreting evidence, and choosing experiments and resource commitments.
The six categories below name the judgment involved and specify what went wrong in a decision.

\begin{itemize}[label={},leftmargin=0pt,labelsep=0pt,topsep=3pt,itemsep=5pt,parsep=0pt]
\item \textbf{Planning.}
The scale, training duration, sequence, or parallelization of experiments does not fit the question that must be resolved next.
This includes both launching a full experiment before checking its pipeline and using a small experiment that cannot test the proposed explanation.
Annotators judge scale against the scientific question, so a larger experiment can be the appropriate correction.

\item \textbf{Adapting.}
A relevant result is already available, but the plan does not use it to revise a hypothesis, configuration, or priority.
The annotator must identify the result that should have changed the next decision.
If the central issue is whether that result is valid, anomaly awareness is the more direct category.

\item \textbf{Anomaly awareness.}
The plan treats an implementation problem, measurement defect, confound, or anomalous output as usable scientific evidence without an adequate diagnostic step.
The issue is whether to check the experiment before interpreting its result, rather than whether the agent can implement the repair.
A correctly executed experiment that answers the wrong research question belongs under focus.

\item \textbf{Evidence sufficiency.}
The available evidence does not support the decision to declare success, failure, completion, or closure.
Examples include rejecting several claims after one immature experiment and treating a noisy result near a threshold as a decisive verdict.
A justified negative conclusion or a stop required by an actual budget limit is not a failure under this category.

\item \textbf{Repetition.}
The proposed work reproduces information already available or presents reused information as independent new evidence.
The annotator must explain what remains scientifically unchanged, such as the predictions, experimental cell, or estimated quantity.
This category excludes planned independent replication and a necessary rerun under a changed protocol.

\item \textbf{Focus.}
Even if executed correctly, the proposed experiment would answer a different question from the one the project is trying to resolve.
This includes substituting a proxy for the target, changing the wrong experimental variable, or treating an agent's own assumption as a binding human constraint.
When a senior researcher introduces a new goal, annotators must mark it as a later change rather than attribute it to the earlier project state.
\end{itemize}

\subsection{Boundaries and Review Procedure}
The primary error label follows the smallest decision change needed to correct the original LLM plan.
Annotators may add up to two secondary error labels when these describe distinct errors rather than restating the same error's cause and consequence.
The six categories have no fixed precedence.
For example, adapting concerns an omitted update, repetition concerns whether proposed work would add information, and evidence sufficiency concerns a conclusion or stop already made.
A case can receive a judgment label without a primary error label.

The project record, an agent's interpretation of it, and the original human message are separate evidence sources.
A project note headed ``human rule'' does not by itself establish that a senior researcher imposed the rule.
Outcomes after a human intervention can explain what happened after the correction, but annotators must not treat them as evidence visible at the decision point.
Related decisions from one human intervention need not be independent instances of an error.

\textbf{Annotation status.}
The historical-error coding assigns a primary error label to 243 cases.
The remaining five cases carry a judgment label without one.
All 248 cases have a judgment label, which names the category of judgment the case concerns.

\subsection{Profiles by Judgment Category}
Figure~\ref{fig:radars}(b) groups all 248 cases by the scientific judgment they concern.
For each model, a spoke shows accuracy in recognizing the reference plan, averaging both option orders within each case and then weighting cases with that label equally.
Categories can overlap, and the scores describe performance on these cases rather than general abilities independent of this set of cases.
Both panels of Figure~\ref{fig:radars} use the same linear 0--100\% radial scale and display all observed values.
The repetition category has only four cases.
Restricting this analysis to the 243 cases with a primary error label changes any model score by at most 1.6 percentage points.
For Figure~\ref{fig:radars}(a), the scientific fields partition the 248 cases, so weighting field scores by case counts recovers each model's overall score.

\subsection{Additional Illustrations}
We selected the examples below for their decision structure, without using the evaluated models' error rates as a selection criterion.
Case numbers refer to the zero-based indices of the benchmark.
Each description summarizes the source material we checked.
Table~\ref{tab:case-observations} reports the corresponding model choices.

\paragraph{A measurement that barely responds to a stiffness change (case 206).}
In a soft-support simulation, a stiffness sweep from $0.25\times$ to $8\times$ changed the center support's static deflection only from roughly 212 to 222\,mm.
All six runs recorded activation of the upper travel limit.
The original LLM plan proposed reducing the number of springs to find the point where visible indentation disappeared and standing performance recovered.
The senior researcher instead asked for an order-of-magnitude estimate of how many springs supported the ball and a check of the effective stiffness and travel limit while retaining the fixed grid.
This case illustrates an anomaly-awareness decision, which is to diagnose whether the simulated response is informative before interpreting the effect of substrate stiffness.
The quoted displacement is the center support's static deflection, not the ball-center displacement.

\paragraph{A gate that delays the main measurement (case 3).}
The original LLM plan kept a $3\times3$ resolution matrix closed until renormalization-group predictions beat the Transferred baseline on both contact treatments under a statistical admission rule.
The measurement that answers the project question is the grid of effective parameters itself, that is, how cloth parameters change with spatial and time resolution.
The senior researcher's note of 2026-08-02 required starting the matrix on the five cross cells.
Under the note, the baseline comparison serves to assess the scale of the error rather than to decide when the matrix may start.
We assign focus because the inherited gate delayed measuring the quantity the project set out to study.

\paragraph{When to shrink an experiment and when to enlarge it (cases 21 and 13).}
For a new batch-API measurement, the original LLM plan took the first reading at 200 items per cell, the size used for the paper tables.
It did so even though an earlier full-size wave had to be discarded after a pipeline bug.
The reference plan ran 20 items per cell on the full grid first and scheduled the 200-item tables after review of those numbers.
In a reinforcement-learning project using proximal policy optimization, the opposite direction was needed.
Three seeds gated at 1,000 iterations could only repeat a known failure with insufficient training and could not distinguish that problem from sensitivity to the random seed.
The reference plan trained one seed for the full 8,000 iterations of the historical protocol before adding more seeds.
Both cases illustrate planning decisions, but the appropriate direction of the scale change is opposite.
The 20-then-200 item protocol and the 8,000-iteration budget are local project requirements, not general optima.

% One sentence per source line. Terminology, style rules, and figure placement: see CLAUDE.md.
\section{Human Baseline Protocol}
\label{app:human-reference}

Fresh researchers with no prior exposure to the source projects answer through an interface that presents the two plans and the associated project record.
They learn the project from that record; none of them supervised the source projects.
Thirteen PhD students and postdocs took part.
Each was informed about the study and gave consent.
All are reported under anonymous identifiers.
The baseline therefore measures what a researcher can recover about the senior researcher's correction from the project record, as the models do.
The interface randomizes option order.
An agent assistant is available to explain the project record and differences between the plans; the fresh researcher makes the final choice.
The default assistant runs \texttt{gpt-5.6-terra} through Codex, with Astra and Claude assistants also available.

\subsection{Coverage and Comparison}
All 248 cases have recorded human responses.
The fresh researchers answer 182 of the 248 cases correctly, an accuracy of 73.4\%.
The radar plots apply this same measure within each scientific field or judgment category.

% One sentence per source line. Terminology, style rules, and figure placement: see CLAUDE.md.
\section{Evaluation Prompts and Available Materials}
\label{app:evaluation-prompts}

\subsection{Twelve Evaluation Conditions}
For each model and case, the evaluation contains ten initial conditions.
These combine the project and options-only conditions with each of the low, medium, high, xhigh, and max requested effort settings.
Two further conditions continue the medium-effort project session through a first and then a second reconsideration.
Each continuation uses the same reconsideration prompt below and retains the preceding conversation, without correctness feedback.
We evaluate every condition in both option orders, giving 24 responses per model and 120 across the five models for each case.

\subsection{Prompts and Available Materials}
The templates below reproduce the evaluation instructions with display wrapping; the Markdown files preserve the original whitespace.
The placeholders \texttt{\{a\}} and \texttt{\{b\}} hold the two plans, swapped for the second option order.
The source archive provides the manuscript, figures, bibliography, prompts, and aggregate model results for all twelve conditions.
The question archive, provided as supplementary material, contains the complete 248-case set with anonymized case identifiers, the two options per case in their exact wording, and reference answers.
Appendix~\ref{app:model-details} documents the evaluation clients and scoring rules, and this appendix provides the prompts.

\noindent\begin{minipage}{\columnwidth}
\paragraph{Project condition.}
\begingroup
\footnotesize
\begin{verbatim}
`/project` is the current state of a research
project. You need to think about what to do
next.

Read idea.md first, then proposal.md, then
STATE.md, and then other files if you think it
is necessary.

Below are two candidate plans. Please choose the
one you think is better.

<option_A>
(A) {a}
</option_A>

<option_B>
(B) {b}
</option_B>

Output format:

```
{
  "argument": "The reason for your choice",
  "response": "A or B"
}
```

Note:

- Only read the contents of `/project`. Do not
go to parent directories. Do not access the
network.
\end{verbatim}
\endgroup
\end{minipage}
\par\medskip

\noindent\begin{minipage}{\columnwidth}
\paragraph{Options-only condition.}
\begingroup
\footnotesize
\begin{verbatim}
Below are two candidate next-step plans for a
research project in a particular state. Please
choose the one you think is better.

<option_A>
(A) {a}
</option_A>

<option_B>
(B) {b}
</option_B>

Output format:

```
{
  "argument": "The reason for your choice",
  "response": "A or B"
}
```
\end{verbatim}
\endgroup
\end{minipage}
\par\medskip

\noindent\begin{minipage}{\columnwidth}
\paragraph{Reconsideration, used for both continuations.}
\begingroup
\footnotesize
\begin{verbatim}
Please reconsider the original question and both
options carefully, then answer again. You may
review the project files if needed. You may keep
or change your previous answer.

Return your answer in the same JSON format as
before:

```json
{
  "argument": "The reason for your choice",
  "response": "A or B"
}
```
\end{verbatim}
\endgroup
\end{minipage}
\par\medskip

\end{document}